\documentstyle[pictexwd,aclap,psfig,11pt]{article}

\title{Chunk Tagger\\ \large Statistical Recognition of Noun Phrases}

\author{Wojciech Skut \and Thorsten Brants\\
        Universit{\"a}t des Saarlandes\\
        Computational Linguistics\\
        D-66041 Saarbr{\"u}cken, Germany\\
        {\tt \{skut,brants\}@coli.uni-sb.de}\\[1ex]
        {\em In ESSLLI-98 Workshop on Automated Acquisition of Syntax
        and Parsing, Saarbr\"ucken, 1998}}

\def\stern{\ifmmode^{\ast}\else$^{\ast}$\fi}
\def\argmax{\mathop{\rm argmax}}
\def\N{\mbox{I\kern-.2emN}}
\def\setgray#1{
\ignorespaces}
\def\unsetgray{
\ignorespaces}

\begin{document}

\maketitle

\begin{abstract}

We describe a stochastic approach to {\em partial parsing},
i.e., the recognition of syntactic structures of limited depth. The
technique utilises Markov Models, but goes beyond usual
bracketing approaches, since it is capable of recognising not only the
boundaries, but also the internal structure and syntactic category of
simple as well as complex NP's, PP's, AP's and adverbials. We compare
tagging accuracy  for different applications and  encoding schemes. 

\end{abstract}


\section{Motivation}

The word {\em chunking} (also {\em partial} or {\em shallow parsing})
refers to techniques used for the recognition of relatively simple
syntactic structures, such as NPs, PPs, verb complexes etc.

NP chunkers typically rely on fairly simple and efficient processing
tools such as finite automata or (in stochastic approaches)
Markov Models (MMs). The output consists of structures recognised with a
high degree of certainty; such structures are used for further
processing.

\subsection{Annotation Software}

The main motivation for the work reported in this paper was the
development of NLP software for creating language resources,
especially syntactically annotated corpora (treebanks). A disadvantage
of symbolic tools supporting corpus annotation is that they are
language specific and have to be developed prior to actual
annotation. For English, this is not a problem since there are many
such tools, yet for other languages, serious difficulties may arise
here.

As an alternative, a bootstrapping approach can be taken in which,
after a short phase of purely manual annotation, more and more
automatic procedures are implemented using statistical NLP methods.
The already annotated sentences serve as training data. This approach
is highly independent of the annotation format, which is simply
learned from training data.

With these prerequisites, we have developed a stochastic parser ({\em
chunker}) that recognises syntactic structures of limited depth. The
tool is language-independent and can be used for parsing unrestricted
text provided some minimal amount of annotated data is available.

\subsection{Overview}

In the following, we describe our stochastic approach to NP chunking
based on a generalisation of standard POS tagging techniques (hence
the name {\em chunk tagger}). First, we show how a simple bracketing
method can be extended to recognise more complex structures and
several types of phrases (sections~\ref{sec:bracketing} and
\ref{sec:cats}). Accuracy for different applications and tasks is 
reported in section~\ref{sec:results}. In section~\ref{sec:related}, we
discuss the similarities and differences between our approach and
related research.

\section{Stochastic NP Recognition}
\label{sec:bracketing}


The idea of using statistics for NP chunking goes back to
\newcite{Church:88}, who used corpus frequencies to determine the 
boundaries of simple non-recursive NP's. For each pair of POS tags
$t_i, t_j$, the probability of an NP boundary (`[' or `]') occurring
between $t_i$ and $t_j$ is computed. On the basis of these context
probabilities, the program inserts the symbols `[' and `]' into
sequences of POS tags, yielding output of the following form:

\bigskip

$[$A/AT former/AP top/NN aide/NN$]$ to/IN $[$Attorney/NP~General/NP~Edwin/NP 
Meese/NP$]$ interceded/VBD to/TO extend/VB $[$an/AT aircraft/NN
company/NN\dots

\bigskip

The accuracy of this approach is impressive. On the other hand, the
task is not too difficult since recursive structures are not
recognised. It is interesting whether this simple technique can be
used for the recognition of more complex phrases.


\subsection{Beyond Simple Bracketing}
\label{sec:chunking}

We have modified Church's approach in a way permitting efficient and
reliable recognition of structures of limited depth, including complex
prenominal adjectival and participial phrases, postnominal PP's and
genitives, appositions, etc. We encode the structure in such a way
that it can be recognised by a part-of-speech tagger, so the process
runs in time linear to the length of the input string.

The basic idea is simple enough: structures of limited depth are
encoded using a finite number of flags. We employ flags standing not
just for bracketing, but for structural relations between adjacent
words.

Given a sequence of words $\langle w_0, w_1, ... w_n\rangle$, we
consider the structural relation $r_i$ holding between $w_i$ and
$w_{i-1}$ for $1 \leq i \leq n$. For the recognition of most -- even
fairly complex -- NPs, PPs, and APs, it is sufficient to distinguish
the following seven values of $r_i$ which uniquely identify
sub-structures of limited depth.

\begin{figure}[h]
\hrule
\bigskip
\begin{center}
\psfig{file=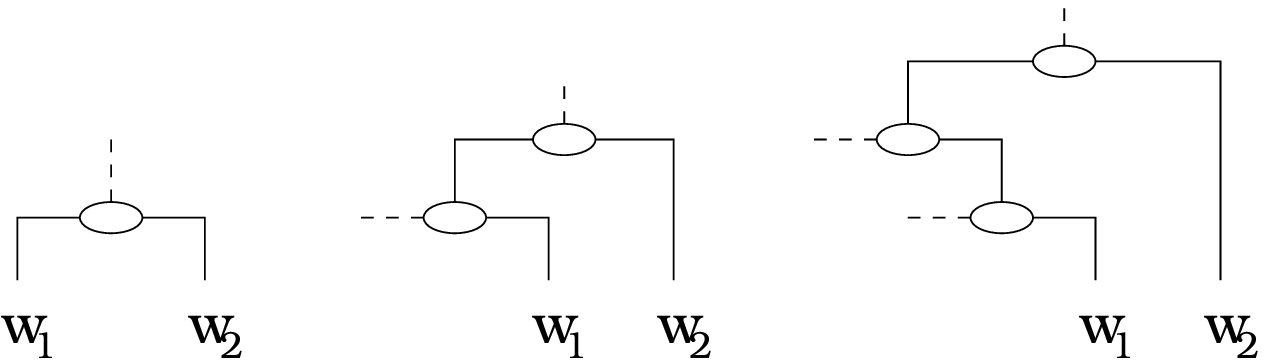,height=1.9cm,rwidth=8cm}
\end{center}
\vspace*{-3ex}
\def\h#1{\hspace*{#1}}
{\sf\h{.6em}$r_2=0$\h{3.5em}$r_2=+$\h{3.8em}$r_2=++$}\\[1ex]
\begin{center}
\psfig{file=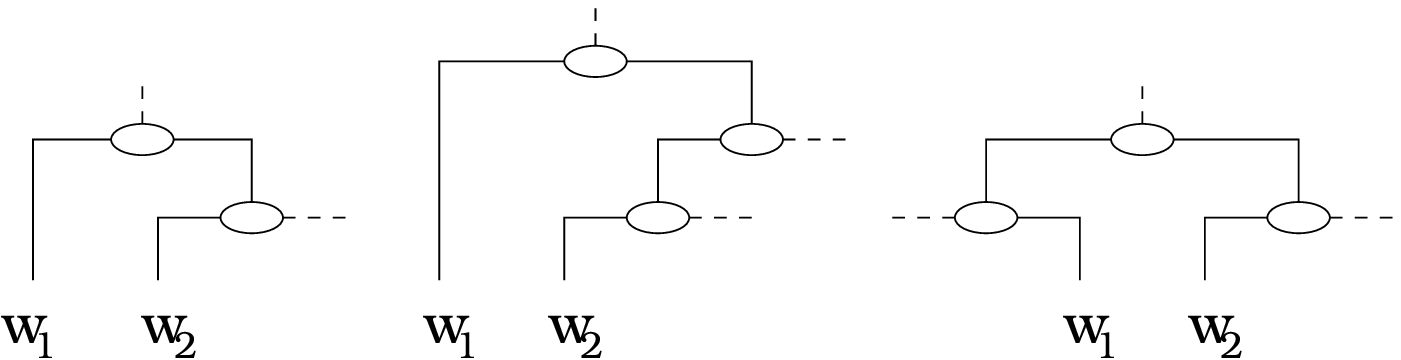,height=1.9cm,rwidth=8cm}
\end{center}
\vspace*{-3ex}
\def\h#1{\hspace*{#1}}
{\sf\h{.6em}$r_2=-$\h{3.5em}$r_2=--$\h{3.8em}$r_2=$ `='}\\[1ex]
\hrule
\caption{Structural tags $r_2$ assigned to $w_2$}
\label{fig:str}
\end{figure}

\begin{figure}[h]
\hrule
\bigskip
\begin{center}
\psfig{file=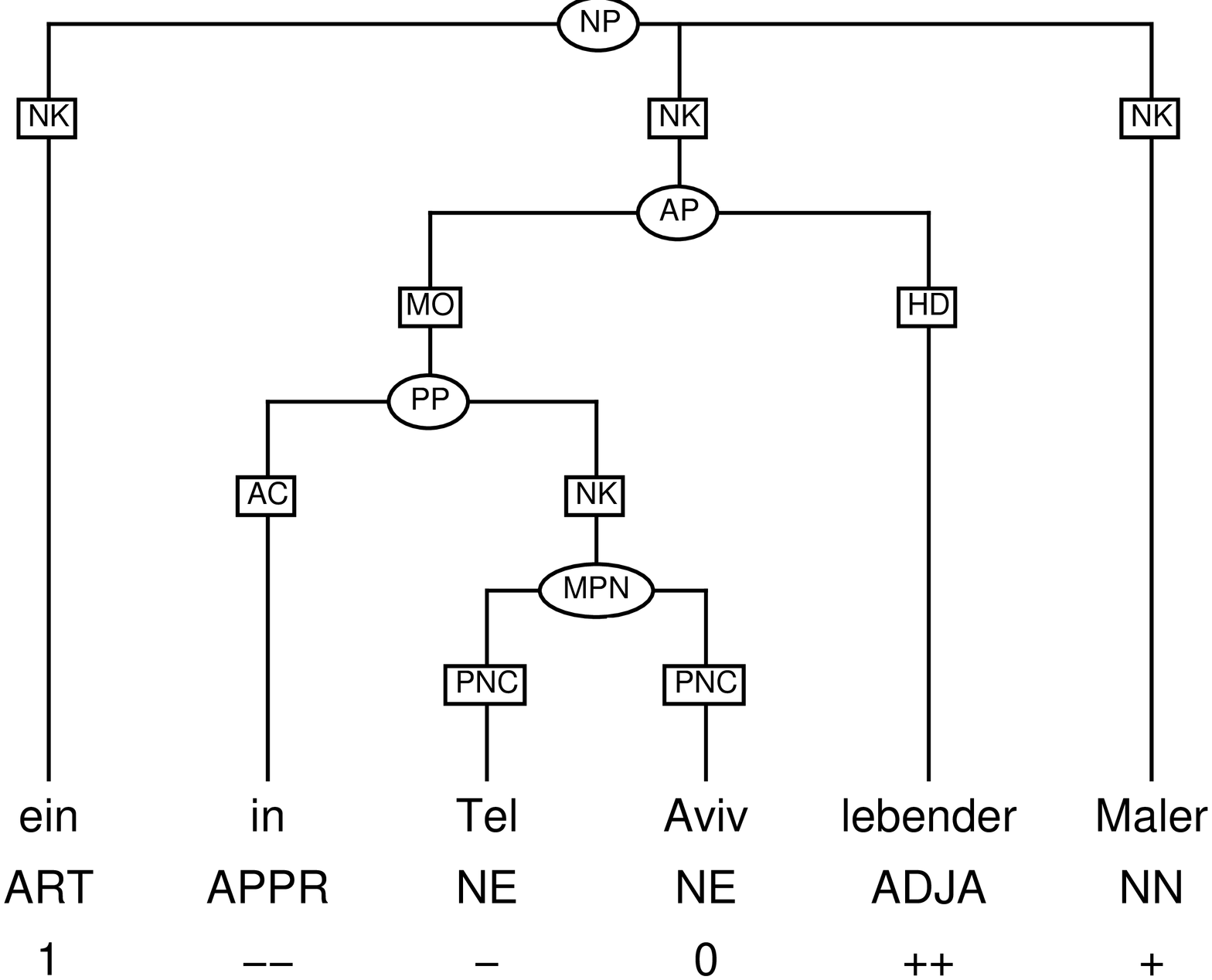,width=8.2cm,rwidth=8cm,angle=-90}
\vspace*{-4ex}
\end{center}
\vspace*{-3ex}
\def\h#1{\hspace*{#1}}
{\sf\h{.6em}a\h{2.9em}in\h{2.4em}Tel\h{2em}Aviv\h{1.5em}living\h{1.3em}painter}\\[1ex]
\centerline{\em `a painter living in Tel Aviv'}

\bigskip

\footnotesize {\sf AC} = adpositional case marker, {\sf HD} = head,
{\sf MO} = modifier, {\sf MPN} = multi-token proper noun,
{\sf NK} = noun phrase kernel, {\sf PNC} = proper noun constituent 

\smallskip

\hrule

\caption{Encoding of a sample structure}
\label{fig:chunks}
\end{figure}

{\tt
\[
r_i = \left \{ \begin{array}{rcl}
        0 & \mbox{\rm if} & parent(w_i)=parent(w_{i-1})\\
        + & \mbox{\rm if} &parent(w_i)=parent^2(w_{i-1})\\
        ++ & \mbox{\rm if} &parent(w_i)=parent^3(w_{i-1})\\
        - & \mbox{\rm if} &parent^2(w_i)=parent(w_{i-1})\\
        -- & \mbox{\rm if} &parent^3(w_i)=parent(w_{i-1})\\
        = & \mbox{\rm if} &parent^2(w_i)=parent^2(w_{i-1})\\
        1 & \mbox{  } & \mbox{\rm else}\\
        \end{array} \right.
\]
}
If more than one of the conditions above are met, the first of the
corresponding tags in the list is assigned.  The depth of structures
is limited to 3. For convenience, we give the graphical representation
of the structural tags in figure \ref{fig:str}. A structure tagged
with these symbols is shown in figure \ref{fig:chunks}.

Instead of the simple context frequencies used by Church, we employ a
generalisation of the standard MM-based POS tagging method. The task
of the chunker is to assign the most probable sequence of structural
tags $R = \langle r_0, r_1, \dots, r_n\rangle$ to a sequence of POS
tags $T = \langle t_0, t_1, \dots, t_n \rangle$. This can be done
exactly in the same way as the assignment of the optimal POS sequence
to a sequence of words in POS tagging, and the task is to calculate
\vbox{%
\begin{eqnarray}
        \argmax_R P(R|T) & & \hspace*{6em}
\end{eqnarray}
\vspace{-2ex}
\begin{eqnarray*}
        &&= \argmax_R \frac{P(R)\cdot P(T|R)}{P(T)}\\[1mm]
        && = \argmax_R P(R)\cdot P(T|R) \\
        && = \argmax_R \prod_{i=1}^k P(r_i|r_{i-2},r_{i-1}) P(t_i|r_i)
\end{eqnarray*}
}

Under this perspective, a standard part-of-speech tagger can be
trained on a syntactically annotated corpus with structures converted
into structural tags (the $r_i$'s). However, in this case the
corresponding Markov Model has only 7 tags (the possible values of $r_i$),
which is obviously too coarse-grained. The precision of the tagger is
rather disappointing: only about 77\% of all structures are recognised
correctly.

To cope with this problem, we enrich the MM state with information
about the POS tag $t_i$ assigned to $w_i$.  Now we can define {\em
structural tags} as pairs $S_i=\langle r_i, t_i\rangle$. Such tags
constitute a finite alphabet of symbols describing structures of depth
$\leq 3$.

The tagger's task is thus to assign the most probable sequence of
structural tags $S= \langle S_0, S_1, ..., S_n\rangle$ to a sequence
of part-of-speech tags $T = \langle t_0, t_1, ..., t_n\rangle$, hence
\vbox{%
\begin{eqnarray}
        \argmax_S P(S|T) & & \hspace*{1em}
\end{eqnarray}
\begin{eqnarray*}
        &&= \argmax_S P(S)\cdot P(T|S)
\end{eqnarray*}
}
The part-of-speech tags are encoded in the structural tag (the $t_i$
dimension), so $S$ uniquely determines $T$. Therefore, we have
$P(t_i|S_i) = 1$ if $S_i = \langle r_i,t_i\rangle$ and 0 otherwise,
which simplifies calculations.

The contexts are smoothed by linear interpolation of unigrams,
bigrams, and trigrams. Their weights are calculated by deleted
interpolation.


\section{Phrasal Categories} 
\label{sec:cats}

A simple extension of the chunk tagger can assign phrasal categories
in addition to structures. We enrich the state $S_i$ of the Markov Model with
information about the category $c_i$ of the node immediately
dominating word $w_i$. Thus $S_i$ becomes a triple $\langle r_i, t_i,
c_i\rangle$. For example, the adjective {\em lebender} in figure
\ref{fig:chunks} is assigned the tag $\langle ++, ADJA, AP\rangle$.
This extension also slightly improves the recognition of structures,
cf. section \ref{sec:results}.

Further precision gain can be achieved if we also add some information
$g_i$ about the category of the grandparent node.  However, only few
symbols can be used to encode this dimension. Otherwise, the tagset
(all $S_i=\langle r_i, t_i, c_i, g_i\rangle$) becomes too large. We
achieved the best results with just three flags $A$, $N$ and $C$,
which indicate that $parent^2(w_i)$ is an AP, an NP/PP and a
coordinated constituent, respectively. In this format, the word {\em
Aviv} in figure \ref{fig:chunks} is assigned the tag $\langle 0,
NE, MPN, N \rangle$.


\section{Applications and Results}
\label{sec:results}

In this section, we compare results achieved for different applications
and types of structures. We use the dependency-oriented NEGRA treebank
\cite{Skut:ea:97a} as training data.
The current size of the corpus is 12,000 sentences, or 210,000 tokens.
All of these sentences have been annotated without the help of the
chunk tagger.

The annotation scheme distinguishes 24 phrasal categories.  The POS
tagset \cite{Thielen:Schiller:94} consists of 54 tags.  For tagging
purposes, it has been adjusted by merging tags irrelevant to the
chunking task and adding simple morphological and lexical
information. Due to this adjustment, 1.5\% more words are assigned the
correct structural tag.

Structures are encoded according to the method presented in
section~\ref{sec:cats}. We vary the number of tag dimensions (1 -- 4).

The results given in the following sections have been computed by spliting
the corpus into disjoint training and test parts (90\% and 10\%,
respectively). This procedure was repeated ten times, and the results
were averaged. The accuracy measures employed are explained as
follows.

\begin{description}
\item[tags:] the percentage of structural tags with the correct value
of the $r_i$ attribute,
\item[bracketing:] the percentage of correctly recognised nodes,
\item[labelled bracketing:] the percentage of nodes  recognised correctly
including their syntactic category,
\item[top-level chunks:] the percentage of correctly parsed ``maximal'' 
chunks, i.e., phrases not contained in a larger chunk of depth
$\leq$ 3. 
\end{description}

We present figures concerning the {\em precision} of the chunker. {\em
Recall} is slightly lower for all applications (0.5\% -- 1.5\%).

\subsection{Corpus Annotation}
\label{sec:annot}

As we already mentioned, the primary application of the chunk tagger
is corpus (treebank) annotation. The treebank is being created in an
interactive annotation mode: automatic and manual annotation steps are
closely interleaved to ensure optimal control of the predictions made
automatically (for a precise description of this interactive approach
to treebank annotation see \cite{Skut:ea:97a}).

As for the chunker, the interactive annotation mode means that the
annotator specifies the boundaries of a complex NP or PP, and the tool
recognises its category and internal structure. Note that the
disambiguation of PP attachment is significantly easier than in the
general case. Correct structural tags are assigned to more than $95\%$
of all words, which means that approx.  $89\%$ of all chunks (NP's,
PP's, AP's) are assigned the correct syntactic structure.

Precise results for different chunk encoding methods are given in
table~\ref{table:res0}. The training corpus was created by extracting
all NPs, PPs and APs occurring in the NEGRA treebank (34,000 chunks,
130,000 tokens). In other words, the program had to perform the
annotator's task: find a labelled structure that spans a given
sequence of words.

\begin{table}[h]
\hrule
\caption{Precision of the chunk tagger in the interactive 
annotation mode for different chunk encoding methods. The symbols in
brackets denote: $r$ structural relation (7 values), $t$ POS tag (54
values), $c$ parent node category (24 values), $g$ grandparent node
category (3 values). }
\label{table:res0}
\bigskip
\begin{center}
\begin{tabular}{l|ccc}
 dimensions                & tags ($r_i$)   & brack.  & l. brack.   \\ \hline
 $\langle r \rangle$       &  87.8\% &  76.6\% & -- \\
 $\langle r,c,g \rangle$   & 92.4\% & 83.9\% & 78.1\% \\
 $\langle r,t \rangle$     & 94.7\% & 88.3\% & --   \\
 $\langle r,t,c \rangle$   & 94.9\% & 88.7\% & 84.7\%  \\
 $\langle r,t,c,g \rangle$ & 95.1\% & 89.2\% &  85.2\% \\
\end{tabular}
\end{center}
\bigskip
\hrule
\end{table}

It can be seen from the table that part-of-speech information ($t$) is
much more relevant than information about phrasal categories ($c$ and
$g$). The latter also enhances the performance of the tagger, but the
improvement achieved is rather modest.

The tagset size and average ambiguity for the five encoding schemes
are shown in table~\ref{table:tagsets}.

\begin{table}[h]
\hrule
\caption{Tagset sizes and ambiguity.}
\label{table:tagsets}
\bigskip
\begin{center}
\begin{tabular}{l|cc}
 dimensions                & \# tags & tags per word  \\ \hline
 $\langle r \rangle$       &  7   &  4.5 \\
 $\langle r,c,g \rangle$   & 125  & 24.9 \\
 $\langle r,t \rangle$     & 251  &  4.5  \\
 $\langle r,t,c \rangle$   & 775 & 18.7 \\
 $\langle r,t,c,g \rangle$ & 996 & 24.9 \\
\end{tabular}
\end{center}
\bigskip
\hrule
\end{table}

With a unigram model, i.e. choosing the most probable tag without
looking at the context, tag assignment precision is only 45.8\% (for
$S=\langle r,t,c,g \rangle$).  The precision of a bigram model is
94.3\%. Thus the difference to the trigram model is very small, which
obviously results from the fairly large amount of information encoded
in a single chunk tag (structural relation, POS tag and phrasal
category), so that a trigram context does not contain much more
information than a bigram one.

\subsection{Tagging the Penn Treebank}

In order to better evaluate the performance of the chunk tagger, we
applied it to a fragment of the Penn Treebank. As in the evaluation
reported in the previous section, the training corpus consisted of all
NP's, PP's and AP's occurring in the Treebank fragment. In the table
below, the results are contrasted with those of chunk tagging the
NEGRA corpus.

\begin{table}[h]
\hrule
\caption{Precision for different corpora in the interactive 
annotation mode (Penn Treebank and NEGRA Corpus). Information
about external phrase boundaries is supplied by the annotator.}
\label{table:res-ptb}
\bigskip
\begin{center}
\begin{tabular}{l|rr}
corpus & PTB & NEGRA \\ \hline
\# sentences & 10,000 & 12,000 \\ 
\# top-level chunks & 33,808  & 33,787 \\ 
\# phrasal nodes & 88,083  & 56,110 \\ \hline
tags ($r_i$) & 93.0\% & 95.1\% \\
bracketing & 91.1\% & 89.2\% \\
lab. bracketing  & 86.7\% & 85.2\% \\
top-level chunks & 81.3\% & 88.8\% \\
\end{tabular}
\end{center}
\bigskip
\hrule
\end{table}

The figures show that the top-level chunk recognition rate is
significantly better for the NEGRA corpus data. The difference seems
to arise from the higher structural complexity of the Penn Treebank
fragment, where a chunk on average contains 2.56 phrasal nodes (as
opposed to 1.65 in the NEGRA corpus, which does not contain unary
projections). 

\subsection{Other Applications}
\label{sec:opti}

The chunk tagger can also be used as a stand-alone application, i.e.,
for the recognition of simple structures in text. This task is
obviously more difficult since all phrase boundaries have to be
recognised by the tagger. As a result, precision drops significantly, 
cf. table~\ref{table:res1}.

\begin{table}[h]
\hrule
\caption{Precision of the chunk tagger with PP/NP/adverb attachment.
No pre-editing by a human  annotator.}
\label{table:res1}
\bigskip
\begin{center}
\begin{tabular}{l|r}
measure & correct \\ \hline
structural tags ($r_i$) & 90.9\% \\
bracketing & 75.4\% \\
labelled bracketing & 72.6\% \\
top-level chunks & 71.1\%
\end{tabular}
\end{center}
\bigskip
\hrule
\end{table}


To find the main sources of errors, we examined the results and found
that erroneous output mostly originated from wrong {\em PP
attachment}. Furthermore, many errors were due to coordination and
{\em focus adverbs} (e.g., {\em nur} `only', {\em auch} `also', etc.),
which introduce a high ambiguity potential.

Since the disambiguation of such attachments involves lexical and even
world knowledge, PP and focus adverb attachment cannot be recognised
in a satisfactory way by a MM-based tagger operating on POS
tags. Thus the best strategy is to postpone attaching PP's and
adverbs, and make the chunk tagger recognise the prenominal part of
NP's and PP's only.  With this modification, precision increases to
more than 95\%. Exact results are given in table \ref{table:res2}.

\begin{table}[h]
\hrule
\caption{Precision of the chunk tagger {\em without} PP/NP/adverb 
attachment. No pre-editing by a human  annotator.}
\label{table:res2}
\bigskip
\begin{center}
\begin{tabular}{l|r}
measure & correct \\ \hline
structural tags ($r_i$) & 95.5\% \\
bracketing & 89.3\% \\
labelled bracketing & 86.2\% \\
top-level chunks & 89.0\%
\end{tabular}
\end{center}
\bigskip
\hrule
\end{table}

If we ignore errors concerning the internal structure of the chunks
(i.e., we measure only the recognition of external boundaries, which
corresponds to the precision measurement in some other approaches),
93.4\% of all chunks are assigned the correct part of the input
string.


\subsection{Size of the Training Corpus}

\begin{figure*}[ht]
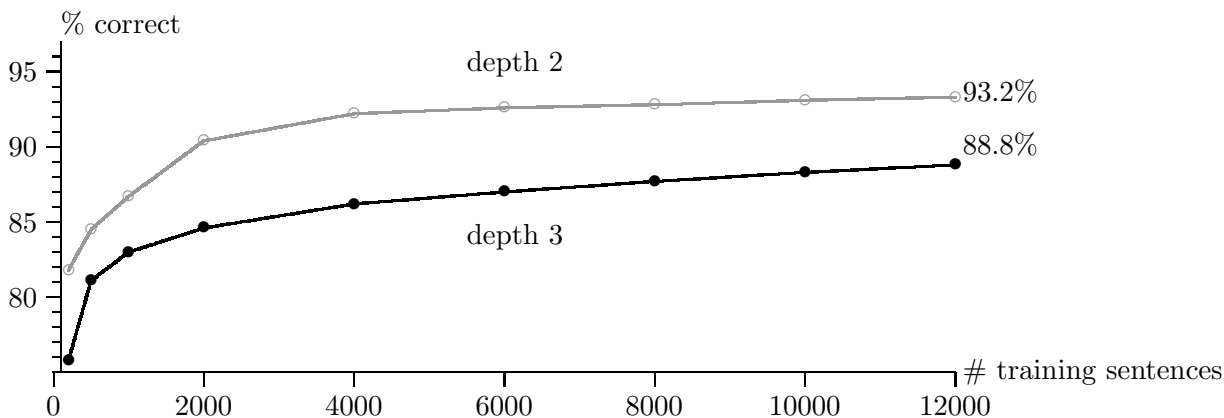

\hrule
\begin{center}
\bigskip
\hspace*{0pt}
\beginpicture
\setcoordinatesystem units <.01mm,2mm>
\setplotarea x from 100 to 12000, y from 75 to 97
\setplotsymbol({\rule{.4pt}{.4pt}})
\axis bottom ticks numbered from 0 to 12000 by 2000 /
\axis left ticks numbered 
        at 80 85 90 95 / /
\axis left ticks short
        from 75 to 96 by 1 /
\put {\# training sentences} [l] at 12100 75
\put {\% correct} [lb] at 95 97.5
\setlinear
\setplotsymbol({\rule{.8pt}{.8pt}})
\plot
200     75.8
500     81.1
1000    83.0
2000    84.6
4000    86.2    
6000    87.0
8000    87.7
10000   88.3
12000   88.8
/
\multiput{$\bullet$} at
200     75.8
500     81.1
1000    83.0
2000    84.6
4000    86.2    
6000    87.0
8000    87.7
10000   88.3
12000   88.8
/
\put {88.8\%} [l] at 12100 90.2
\put {depth 3} [l] at 5500 84

\setgray{.6}
\setplotsymbol({\rule{.8pt}{.8pt}})
\plot
200     81.8
500     84.5
1000    86.7
2000    90.4
4000    92.2
6000    92.6
8000    92.8
10000   93.1
12000   93.3
/
\multiput{$\circ$} at
200     81.8
500     84.5
1000    86.7
2000    90.4
4000    92.2
6000    92.6
8000    92.8
10000   93.1
12000   93.3
/
\unsetgray

\put {93.2\%} [l] at 12100 93.7
\put {depth 2} [l] at 5500 95.5

\endpicture
\medskip
\end{center}
\hrule
\caption{Precision as percentage of correctly recognised top-level 
chunks of depth 2 and 3, shown for different numbers of training sentences.}
\label{fig:results}
\end{figure*}

An important advantage of the chunker is that it is independent of
theory-internal representations and can be used to recognise
structures of any language. Of course, the availability of a training
corpus is a prerequisite. Now we shall see how much data is necessary
to achieve reliable results.

Figure~\ref{fig:results} shows precision (measured as the percentage
of top-level chunks recognised correctly) for the interactive
annotation mode. We consider two encoding schemes. The {\em depth 3}
scheme is the one described in section~\ref{sec:chunking}, which uses
all the 7 possible values of the $r_i$ dimension. The {\em depth 2}
scheme is its slightly simplified version in which $r_i$ can take only
four values: {\bf 1, 0, +, -}, so that only depth-two trees are
recognised by the chunker.

While for the depth 3 encoding a training corpus of 1000--2000
sentences is needed, the simpler encoding requires only about 500
sentences. Thus the chunk tagger can be successfully used in treebank
annotation at quite an early stage, with only a few hundred annotated
sentences available.

\section{Related Work}
\label{sec:related}

In section \ref{sec:bracketing}, we sketched the simple bracketing
technique described by \newcite{Church:88}, which provided motivation for
our chunking method. As far as other approaches are concerned, our
work is most closely related to that of \newcite{Joshi:Srinivas:94}, who
use Markov Models in a preprocessing step to reduce the number of tree segments
(called {\em supertags}) that can be assigned to a word in a
lexicalised Tree Adjoining Grammar. 
This approach makes parsing more efficient, but it needs a large
training corpus, has to fight a large amount of ambiguity and needs a
subsequent parsing step (also see \cite{Srinivas:96} for the use of
explanation-based learning for this purpose).

Symbolic NP chunkers usually rely on finite automata and/or pattern
matching, cf. \cite{Koskenniemi:90},
\cite{AitMokhtar:Chanod:97}. \newcite{Abney:96} presents a partial
parsing technique based on cascaded finite
automata. \newcite{Voutilainen:Padro:97} describe a POS tagger and
shallow parser combining symbolic and stochastic processing via {\em
relaxation labelling}. 

The precision of the abovementioned approaches is often measured by
the number of correct labels assigned to words. The figures range from
85\% to 98\%.  Our results (89\% -- 95\%) fit into this interval, yet
it should be kept in mind that the coverage of the approaches and the
precision measuring methods are often only roughly comparable.  For
instance, several shallow parsing methods are restricted to POS
tagging and grammatical function labelling without explicitly
specifying attachments and phrase boundaries.  Furthermore, the notion
of `phrase' varies in these investigations, and usually these studies
concentrate on simple, non-recursive structures. By contrast, our
chunker is capable of recognizing complex, even recursive, NPs, PPs,
and APs.

Compared to the symbolic techniques, an important advantage of the
stochastic approach taken in our project is its independence of
external lexical resources. As a result, the chunker trained with the
POS-tags and structures of the current corpus is fairly
domain-independent. Of course, our tool would benefit from the use of
lexical knowledge; this issue has to be addressed in the near future.

Since our approach is restricted to a small number of structurally
different tags, it has the great advantage of requiring only a small
amount of training data (cf.\ section \ref{sec:results}) and the
recognition of these phrases is of high accuracy.


\section{Conclusion}

We have presented a stochastic partial parser ({\em chunker}) that
recognises the boundaries, internal structure and syntactic category of
simple as well as fairly complex NP's, PP's and AP's. The chunker is a
straightforward application of a stochastic part-of-speech tagger. We
use it to model a mapping from lexical categories to syntactic
structures. The type of the structural encoding is crucial in this
approach, and better encodings increase the accuracy of structural
assignment. The use of Markov Model processing techniques guarantees that the
process runs in time linear to the length of the input string.


\end{document}